\documentclass[onecolumn]{IEEEtran}

\usepackage{fancyhdr}
\IEEEoverridecommandlockouts
\usepackage{orcidlink}
\usepackage{hyperref}
\usepackage{amssymb,amsfonts}
\usepackage{algorithm}
\usepackage[noend]{algpseudocode}
\usepackage{graphicx}
\usepackage{textcomp}
\usepackage{xcolor}
\usepackage{bm}
\usepackage{mathtools}
\usepackage{url}
\usepackage{graphicx}
\usepackage[bottom]{footmisc}
\usepackage{amsmath}
\usepackage{ctable}
\usepackage{cuted}
\usepackage{graphics}
\usepackage{csquotes}
\usepackage{enumitem}

\newtheorem{definition}{Definition}

\newtheorem{example}{Example}

\usepackage{caption}
\def\BibTeX{{\rm B\kern-.05em{\sc i\kern-.025em b}\kern-.08em
    T\kern-.1667em\lower.7ex\hbox{E}\kern-.125emX}}

\begin{document}

\title{Learning a Sparse Neural Network using IHT\\
}

\author{\IEEEauthorblockN{Saeed Damadi\textsuperscript{1} \orcidlink{0000-0002-2806-1476}, 
Soroush Zolfaghari\textsuperscript{2}, 
Mahdi Rezaie\textsuperscript{3},
Jinglai Shen\textsuperscript{4}\\}
\IEEEauthorblockA{
\textit{\textsuperscript{1,4}Department of Mathematics and Statistics} \\
\textit{\textsuperscript{2,3}Department of Computer Engineering} \\
\textit{\textsuperscript{1,4}University of Maryland, Baltimore County (UMBC), Baltimore, USA} \\
\textit{\textsuperscript{2}Isfahan University of Technology, Isfahan, Iran} \\
\textit{\textsuperscript{3}Amirkabir University of Technology, Tehran, Iran} \\
Email: \textsuperscript{1,4}\{sdamadi1, shenj\}@umbc.edu, \textsuperscript{2}zolfaghari.soroush@gmail.com, \textsuperscript{3}mahdi.rezaie.336@aut.ac.ir}
}

\maketitle

\begin{abstract}

The core of a good model is in its ability to focus only on important information that reflects the basic patterns and consistencies, thus pulling out a clear, noise-free signal from the dataset. This necessitates using a simplified model defined by fewer parameters. The importance of theoretical foundations becomes clear in this context, as this paper relies on established results from the domain of advanced sparse optimization, particularly those addressing nonlinear differentiable functions. The need for such theoretical foundations is further highlighted by the trend that as computational power for training NNs increases, so does the complexity of the models in terms of a higher number of parameters. 
In practical scenarios, these large models are often simplified to more manageable versions with fewer parameters.

Understanding why these simplified models with less number of parameters remain effective raises a crucial question. Understanding why these simplified models with fewer parameters remain effective raises an important question. This leads to the broader question of whether there is a theoretical framework that can clearly explain these empirical observations. 
Recent developments, such as establishing necessary conditions for the convergence of iterative hard thresholding (IHT) to a sparse local minimum—a sparse method analogous to gradient descent, are promising. The IHT algorithm's remarkable capacity to accurately identify and learn the locations of nonzero parameters underscores its practical effectiveness and utility.

This paper aims to investigate whether the theoretical prerequisites for such convergence are applicable in the realm of neural network (NN) training by providing justification for all the necessary conditions for convergence. Then, these conditions are validated by experiments on a single-layer NN, using the IRIS dataset as a testbed. Our empirical results demonstrate that convergence conditions can be reliably ensured during the training of a neural network (NN), and under these conditions, the IHT algorithm consistently converges to a sparse local minimizer.

\end{abstract}

\begin{figure}[t]
    \centering 
    \includegraphics[scale=0.6]{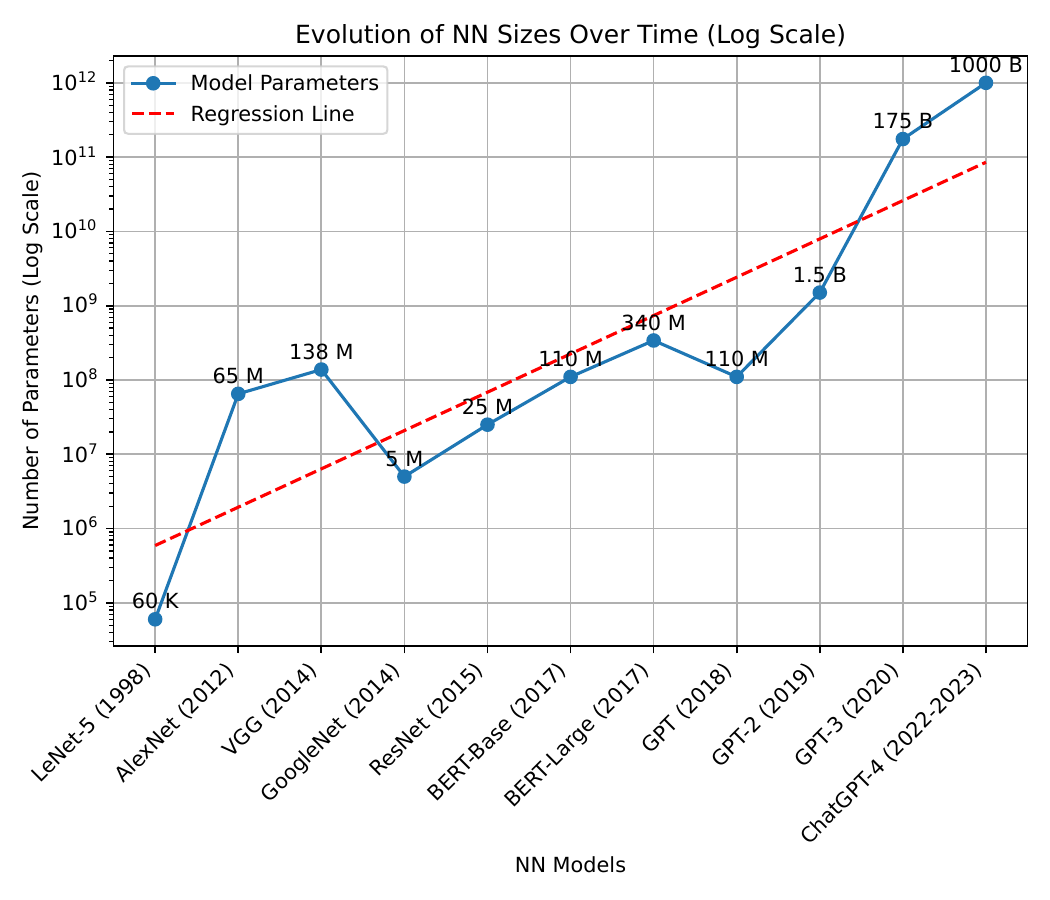}
    \caption{Evolution of NN sizes over time}
    \label{fig:growth}
\end{figure}

\section{Introduction}
Machine learning models, often overparameterized, have the capability to completely fit the training data. In simple terms, larger models tend to memorize all the samples.
In information theory terms, overfitting is described as a scenario where a model's capacity far exceeds the actual informative content of the data it's trained on. This leads to the model not only learning useful patterns and regularities but also absorbing noise and irrelevant details, which reduces its ability to generalize well to new data.
From the perspective of optimization, this
implies achieving a global minimum in the training phase.
The model should ideally capture only the essential information that reflects the underlying pattern or structure. In essence, the goal is to extract the dataset's noise-free signal using a model that is as simple as possible.
To achieve this goal, the optimization problem is formulated as follows: 
\begin{equation}\label{eq:sparseoptimization}
\quad
\begin{array}{l}
\min f(\bm{\theta}, \mathbf{X}, \mathbf{Y}) 
\text{ s.t. }
\bm{\theta} \in \mathbb{R}^n \text{ with }
\|\bm{\theta}\|_0 \leq s,
\end{array}
\end{equation}
where $\bm{\theta}$ is the vector of parameters, $\mathbf{X}$ represents the input data, $\mathbf{Y}$ the output data, given $s$ (with$1 \leq s<n$) denotes the sparsity level, and $\|\cdot\|_0$ is the zero norm, defined as the count of nonzero elements in a vector.
To elaborate on (\ref{eq:sparseoptimization}), recall that the conventional training phase of deep NNs finds $\bm{\theta}^*$ as a local minimizer of the following problem:
\begin{equation}\label{eq:denseoptimization}
\quad
\begin{array}{l}
\min f(\bm{\theta}, \mathbf{X}, \mathbf{Y}) 
\text{ s.t. }
\bm{\theta} \in \mathbb{R}^n.
\end{array}
\end{equation}
To approximately solve the sparse optimization problem in (\ref{eq:sparseoptimization}), one needs to find $\bm{\theta}_s^*$ as a local minimizer of (\ref{eq:sparseoptimization}) with $\|\bm{\theta}_s^*\|_0 \leq s$ such that the following “$\epsilon$-optimality condition” holds for sufficiently small $\epsilon>0$, i.e., 
\begin{equation}\label{eq:epsopt}
\min f(\bm{\theta}^*, \mathbf{X}, \mathbf{Y}) 
+ \epsilon
\leq
\min f(\bm{\theta}_s^*, \mathbf{X}, \mathbf{Y}),    
\end{equation}
where $\bm{\theta}_s^*$ represents parameters of a trained sparse network that has at most $s$ nonzero elements. Consequently, this sparse network achieves a loss value close to that of the dense one. Hence, the trained sparse model not only consumes less energy but also requires less hardware memory and performs inferences faster \cite{damadi2023amenable}.

Reducing the number of parameters in a NN is not a new endeavor, tracing back over thirty years to LeCun's 1989 work \cite{lecun1989optimal}. The need for models with fewer parameters emerges when the largest models, which capture more nuances and patterns with higher accuracy, are constrained by computational complexity. The Compound Annual Growth Rate (CAGR) of parameters in NN models from 1998 (starting with LeNet-5 \cite{lecun1998gradient}) to ChatGPT-3 is approximately 97\% per year. This growth, as evident from the logarithmic scale in Fig. \ref{fig:growth}, follows
a linear trend. The 97\% CAGR per year means that, on average, the number of parameters in these NN models nearly doubled every year between 1998 and 2020. By 2040, the projected number of parameters in a NN model, based on the current growth rate, would be approximately $1.32\times10^{17}=132\times10^{15}$ or 132 quadrillion. The projected number of parameters in NN models by the year 2040 (132 quadrillion) is about 1,320 times greater than the estimated 100 trillion synaptic connections in the human brain. This comparison highlights the immense scale at which NN models are growing, potentially surpassing the complexity of connections found in the human brain by a significant margin. Although 132 quadrillion parameters is deemed to be a large number, it is approximately 3,500 times larger than the number of cells in the human body, which is 37.2 trillion. Hopefully, by 2040 we will have NNs capable of extracting features representing regularities among human body cells. Training such immense models in the future will require a profound understanding of these models in order to simplify them. 
This understanding is crucial for scaling up to networks that are sufficiently large.

The most straightforward way of this simplification would be sparsification of these models that result in fewer parameters. In other words, finding a sparse model within the dense structure. However, the question is whether such a sparse network exists in a dense structure or not.
To address this question, the Lottery Ticket Hypothesis \cite{frankle2018lottery} posits that 
\textquote{within a large, randomly initialized network, there exist smaller subnetworks (termed `winning lottery tickets') that, when trained in isolation from the beginning, can reach similar performance as the original network in a comparable number of iterations}.
This hypothesis suggests that these smaller networks are capable of efficient training. The idea challenges the common practice of using large networks and encourages finding and training these smaller, more efficient subnetworks for similar performance outcomes.

\begin{center}
\begin{minipage}{0.5\linewidth} 
\begin{algorithm}[H] 
\caption{The iterative hard thresholding (IHT)}
\label{alg:IHT}
\begin{algorithmic}[1]
\Require $\bm{\theta}^0 \in \mathbb{R}^n$ such that $\|\bm{\theta}^0\|_0 \leq s$ and stepsize $\gamma > 0$.
\State $\bm{\theta}^{k+1} \in H_s(\bm{\theta}^k - \gamma \nabla f(\bm{\theta}^k))$ for $k=0,1,\dots$
\end{algorithmic}
\end{algorithm}
\end{minipage}
\end{center}

\begin{center}
\begin{minipage}{0.38\linewidth} 
\begin{algorithm}[H] 
\caption{The Gradient Descent Algorithm}
\label{alg:GD}
\begin{algorithmic}[1]
\Require $\bm{\theta}^0 \in \mathbb{R}^n$ and stepsize $\gamma > 0$.
\State $\bm{\theta}^{k+1} = \bm{\theta}^k - \gamma \nabla f(\bm{\theta}^k)$ for $k=0,1,\dots$
\end{algorithmic}
\end{algorithm}
\end{minipage}
\end{center}


Following this direction, there is theoretical evidence in advanced sparse optimization that increases the chance of correctness of this hypothesis.
The theoretical results found by Damadi and Shen in \cite{damadi2022gradient} lay out the theory of finding a sparse local minimizer of a general objective function, of which (\ref{eq:sparseoptimization}) is a special case. They prove that the Iterative Hard Thresholding
\footnote{The Hard Thresholding operator in Alg. \ref{alg:IHT}, $H_s(\cdot)$ keeps the $s$ largest entries of its input in absolute values sense, e.g., $H_2([1,-3,1]^{\top})$ is either $[0,-3,1]^{\top}$ or $[1,-3,0]^{\top}$.}
algorithm (IHT) sequence converges to a HT-stable stationary point. The IHT algorithm uses the full gradient which can be calculated efficiently using the Backpropagation algorithm \cite{damadi2023backpropagation}. The stochastic version of the IHT algorithm known as the SIHT has already been addressed in \cite{damadi2023convergence} which is beyond the scope of this paper. Note that the IHT algorithm illustrated in Alg. \ref{alg:IHT} is the sparse analogue of the gradient descent algorithm outlined in Alg. \ref{alg:GD}. Similarly, the HT-stable stationary point is the analogue of a local minimum in conventional optimization. The results in \cite{damadi2022gradient} guarantee the convergence under certain conditions, while an arbitrary step-length (learning rate) in $(0, 1/L_s]$ and any arbitrary sparse initialization is allowed.

As results in \cite{damadi2022gradient} address a general sparse setting under certain conditions, one needs to examine whether these theoretical results can be applied to NNs or not. To do so, one needs to answer a couple of questions:
\begin{enumerate}[label=Q\arabic*)]
\item What are core assumptions about the objective function in \cite[Optimization problem (1)]{damadi2022gradient}?
\item 
Is it possible to ensure that assumptions posited in \cite[
Corollary 5]{damadi2022gradient} remain valid during the training of a NN?
\item 
Suppose the above assumptions hold. In theoretical results, certain parameters are used. The question then arises: when simulating, how can these parameters be estimated? This is essential for their utilization during simulation.
\item 
Assuming the availability of an estimator for the parameters used in \cite[
Corollary 5]{damadi2022gradient}, does the IHT sequence generated by the   Alg. \ref{alg:GD}, demonstrate numerical convergence to an HT-stable point?
\end{enumerate}

In this paper, we try to answer the above questions and verify the theoretical results of \cite{damadi2022gradient} experimentally. Developing theoretical results requires strong assumptions that guarantee those results. Thus, it increases the computational complexity of experiments. That is why we address a small-sized network like a single-layer NN trained on the IRIS dataset. Note that the purpose of this paper is to spot the IHT algorithm as a valid candidate for simplifying NNs. This is a worthwhile goal because the main success of deep NNs lies in their capability for self-correction, which is facilitated by the gradient descent type algorithm, invented in the 19th century by Cauchy \cite{cauchy1847methode}.

\begin{figure}[t]
    \centering  \includegraphics[scale=0.118]{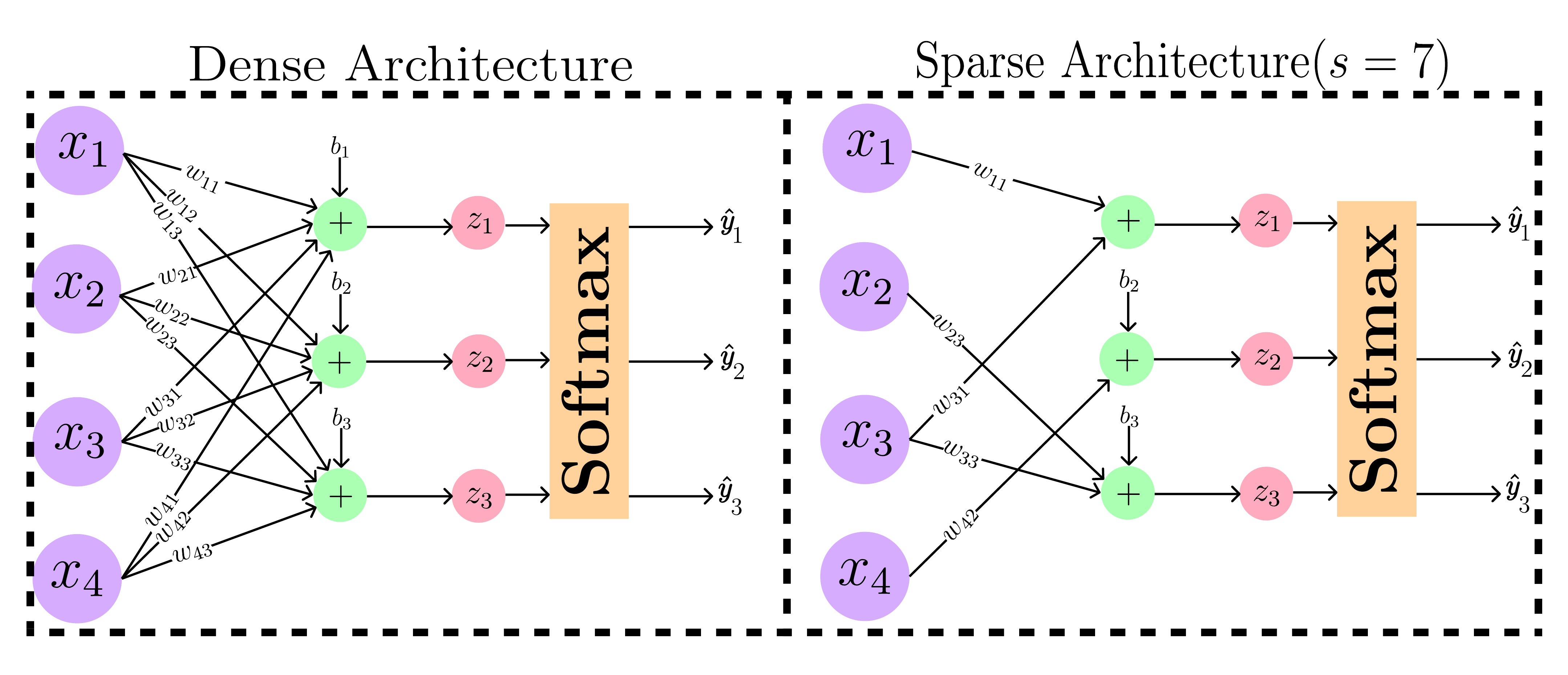}
    \caption{A dense architecture vs a spare one}
\label{fig:densevssparse}
\end{figure}
\section{Related Work}

Sparsification has been extensively examined, with \cite{hoefler2021sparsity} presenting a comprehensive overview of various methods. These approaches are generally classified into two categories: `data-free' and `training-aware'. Data-free methods, such as magnitude pruning by \cite{han2015deep} and \cite{strom1997sparse}, focus on pruning models based on NN structures, using the absolute value of parameters as an importance indicator. On the other hand, training-aware sparsification uses data to determine parameters with minimal impact on the output. This approach includes methods based on first-order approximations, as seen in \cite{xiao2019autoprune,ding2019centripetal}, and second-order approximations, such as those by \cite{hassibi1993optimal,lecun1989optimal}. Variational Dropout (VD) employs a unique approach, modeling parameter distributions and eliminating those with significant noise \cite{molchanov2017variational}. Surprisingly, \cite{gale2019state} conducted a broad comparison of sparsification methods, finding that the simpler magnitude pruning can be as effective or even superior to the more complex VD in large models. Despite the unanimous agreement about the magnitude pruning in the literature, the theory behind it has not been fully addressed, which will be addressed in this paper. For the reader's information, Neuralmagic \cite{neuralmagic_sparsezoo} provides a range of sparsified large-scale NNs.

\section{The IHT algorithm convergence setting}

In this section, we explore the convergence properties of the Iterative Hard Thresholding (IHT) algorithm, a key technique in sparse NN optimization. This section focuses on the critical theoretical foundations and conditions necessary for the algorithm's effective convergence, emphasizing its unique role in simplifying complex NN models.

When Algorithm \ref{alg:IHT} is executed, the objective is to find a local minimizer of the optimization problem in (\ref{eq:sparseoptimization}) that satisfies the ``$\epsilon$-optimality condition" as described in (\ref{eq:epsopt}). Specifically, this means the function value $f$ at $\bm{\theta}^*$ should be within an $\epsilon$ range of its minimum value at $\bm{\theta}_s^*$. If achieved, this would yield an affirmative response to Q4, aligning with the primary aim of this paper. One can visualize $f(\bm{\theta}^*, \mathbf{X}, \mathbf{Y})$ and $f(\bm{\theta}_s^*, \mathbf{X}, \mathbf{Y})$ in (\ref{eq:epsopt}) as the loss functions of two trained networks, where the former corresponds to a dense network and the latter to a sparse network depicted in Fig. \ref{fig:densevssparse}. It is important to note that the nonzero parameters of $\bm{\theta}_s^*$, such as $w_{11}, w_{31}, w_{42}, w_{23}, w_{33}, b_2$, and  $b_3$, are only determined after the Iterative Hard Thresholding (IHT) algorithm is used. The strength of the IHT algorithm is that it learns the parameters and the mask at the same time. This approach not only refines the model's parameters but also sets the stage for examining the underlying assumptions of the objective function, which is crucial for our subsequent analysis.

Turning our attention to addressing questions Q1-Q4, we logically start with Q1 which is about properties of the objective function. The primary assumption regarding the objective function, as discussed in \cite{damadi2022gradient}, is its Restricted Strong Smoothness (RSS) property, outlined in \cite[Definition 1]{damadi2022gradient}. In the next section, we will examine this property more closely by comparing it to the concept of Strong Smoothness. Comprehending this concept is vital as it lays the foundation for answering Q2 and facilitates the transition from theoretical findings to their application in NNs. Other assumptions in \cite[Corollary 5]{damadi2022gradient} include the boundedness of every sublevel set over each subspace with a dimension of at most $s$, and distinct objective values for each distinct HT-stable stationary point.

\subsection{Restricted Strong Smoothness vs Strong Smoothness}
To ensure completeness, the definitions of restricted SS and SS are provided here.
\begin{definition}\label{def:rss}
A differentiable function $f: \mathbb{R}^n \to \mathbb{R}$ is said to be Restricted Strongly Smooth (RSS) with modulus $L_{2s}>0$ or is $L_{2s}$-RSS if
\begin{equation}\label{eq:rss}
f(\tilde{\bm{\theta}}) \leq f(\bm{\theta}) + \langle \nabla f(\bm{\theta}) , \tilde{\bm{\theta}}-\bm{\theta} \rangle + \frac{L_{2s}}{2}\|\tilde{\bm{\theta}}-\bm{\theta}\|_2^2
\end{equation}
holds for all $\bm{\theta},\tilde{\bm{\theta}} \in \mathbb{R}^n$ such that $\|\tilde{\bm{\theta}}\|_0 \leq s$ and $\|\bm{\theta}\|_0\leq s$.
\end{definition}
Recall that the so-called Strong Smoothness property is much stronger than RSS because it requires a global second-order upper bound on a function as follows:
\begin{definition}\label{def:ss}
A differentiable function $f: \mathbb{R}^n \to \mathbb{R}$ is said to be Strongly Smooth (SS) with modulus $L>0$ or is $L$-SS if
\begin{equation}\label{eq:ss}
f(\tilde{\bm{\theta}}) \leq f(\bm{\theta}) + \langle \nabla f(\bm{\theta}) , \tilde{\bm{\theta}}-\bm{\theta} \rangle + \frac{L}{2}\|\tilde{\bm{\theta}}-\bm{\theta}\|_2^2
\end{equation}
holds for all $\bm{\theta},\tilde{\bm{\theta}} \in \mathbb{R}^n$.
\end{definition}
\subsection{Sufficient Condition for Being RSS or SS}\label{sub: sufficient}

It is well-known that when the gradient of a differentiable function $f: \mathbb{R}^n \rightarrow \mathbb{R}$ is Lipschitz with a constant $L>0$, the function is also SS with the same constant $L$. That is, $f: \mathbb{R}^n \rightarrow \mathbb{R}$ is SS with an $L>0$ if

\begin{equation}\label{eq:lg}
\| \nabla f(\tilde{\bm{\theta}}) - \nabla f(\bm{\theta}) \|_2
\leq L\|\tilde{\bm{\theta}}-\bm{\theta}\|_2
\end{equation}

for any $\tilde{\bm{\theta}}, \bm{\theta} \in \mathbb{R}^n$. A similar result can be established for the sparse case. Specifically, if the gradient of $f$ is Lipschitz with a constant $L_{2s}$ for all $\bm{\theta},\tilde{\bm{\theta}} \in \mathbb{R}^n$ satisfying $\|\tilde{\bm{\theta}}\|_0 \leq s$ and $\|\bm{\theta}\|_0\leq s$, then $f$ is $L_{2s}$-RSS.

\subsection{RSS and SS Interpretations and Connections}

There are three main differences between the RSS and SS properties as follows:
\begin{enumerate}
\item
The modulus of the RSS property ($L_{2s}$) depends on the sparsity level, i.e., $L_{2s}$ is a function of $s$. On the other hand, $L$ is a global positive constant for the SS property.
\item
The SS property is valid over the entire $\mathbb{R}^n$, but the RSS property must be applicable over each subspace of $\mathbb{R}^n$ with a dimension of at most $2s$. In other words, there is a second-order upper bound on the function $f$, restricted to some coordinates forming subspaces with a dimension of at most $2s$.
\item
It is always the case that $L_{2s} \leq L$, as $L$ is a global constant \cite{beck2013sparsity}.
\end{enumerate}

\begin{figure}[t]
    \centering
    \includegraphics[scale=0.2]{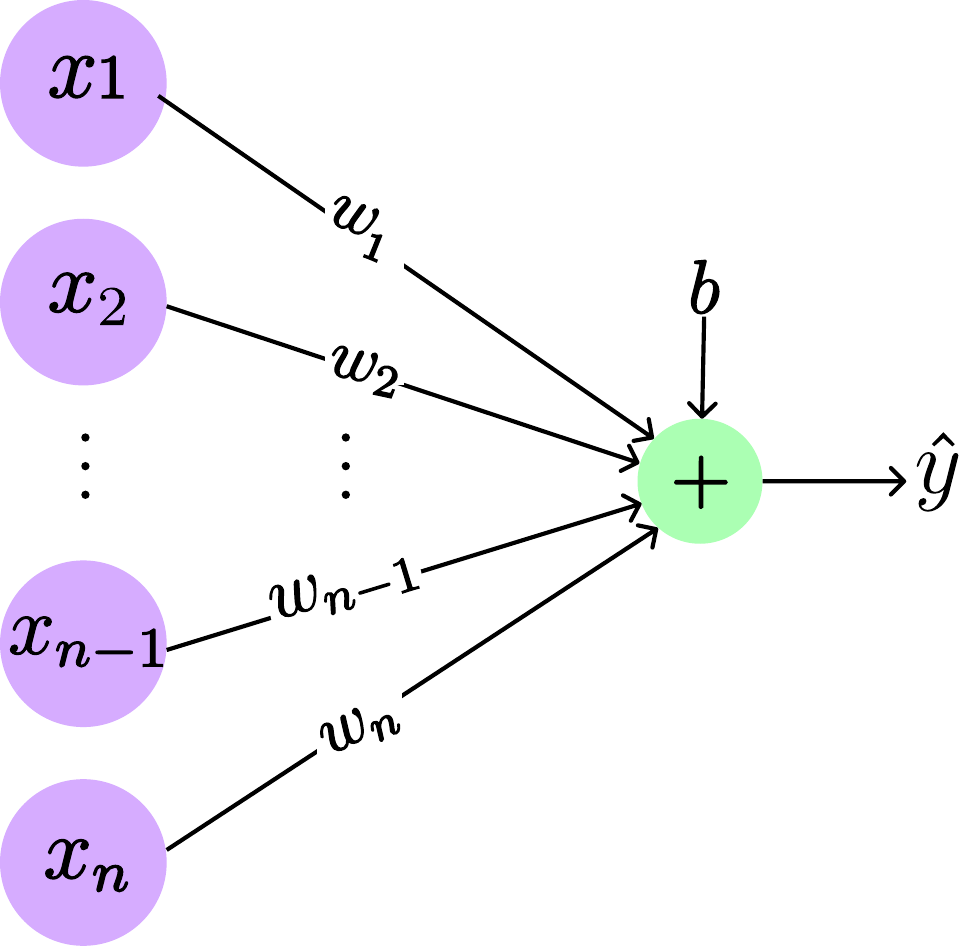}
    \caption{One-Layer NN with $\bm{\theta}=[w_1,\dots, w_n, b]^{\top}$.}
    \label{fig:one-layer}
\end{figure}

The following example differentiates between the RSS and SS properties using a quadratic loss function. This example aims to clarify the concept while avoiding the technicalities that might arise in a more complicated function.

\begin{example}\label{eg:ssandrss}
Let $f: \mathbb{R}^{n+1} \to \mathbb{R}$ where 

\begin{equation}\label{eq:quad}
f(\bm{\theta})=\frac{1}{2}||\mathbf{X}\bm{\theta}-\mathbf{y}||^2,  
\end{equation}
$\mathbf{X} \in \mathbb{R}^{m \times (n+1)}$ and $\mathbf{y} \in \mathbb{R}^m$. As Fig. \ref{fig:one-layer} illustrates, this function represents the overall loss function of a one-layer NN with a mean squared error loss function, whose vector of parameters is $\bm{\theta}=[w_1,\dots, w_n, b]^{\top}$.  
It can be observed that the left-hand side of (\ref{eq:lg}) for this function is $\|\mathbf{X}^{\top}\mathbf{X}(\bm{\theta}-\tilde{\bm{\theta}})\|$. If $\bm{\theta}$ and $\tilde{\bm{\theta}}$ can take any values in $\mathbb{R}^{n+1}$, then the following holds:
$$
\|\mathbf{X}^{\top}\mathbf{X}(\bm{\theta}-\tilde{\bm{\theta}})\|_2\leq \|\mathbf{X}^{\top}\mathbf{X}\|\|\bm{\theta}-\tilde{\bm{\theta}}\|_2  
\quad \forall \bm{\theta},\tilde{\bm{\theta}} \in \mathbb{R}^{n+1},
$$
which implies $L=\sqrt{\lambda_{\max}((\mathbf{X}^{\top}\mathbf{X})^2)}=\lambda_{\max}(\mathbf{X}^{\top}\mathbf{X})$ where $\|\cdot\|_2$ is Euclidean norm. If $n+1 \leq m$ (the number of data points is at least equal to the number of parameters), then there always exists an $L>0$ and the function becomes SS. It should be noted that if $n+1 < m$, (\ref{eq:lg}) does not hold and the sufficient condition is not met. Hence, one needs to check the condition in (\ref{eq:ss}). As a result, the function may or may not be SS. On the other hand,
when
$\bm{\theta}$ and $\tilde{\bm{\theta}}$ take values in $\mathbb{R}^{n+1}$ with $\|\bm{\theta}\|_0\leq s$ and $\|\tilde{\bm{\theta}}\|_0\leq s$, then $L_{2s}=\lambda_{\max}(\mathbf{X}_{\bullet\mathcal{S}}^{\top}\mathbf{X}_{\bullet\mathcal{S}})$ where $\mathbf{X}_{\bullet\mathcal{S}}$ is the restricted columns of $\mathbf{X}$ associated with coordinates of $\bm{\theta}$ and $\tilde{\bm{\theta}}$ that are both nonzero, i.e., 
$$
\mathcal{S}:=\{i \bigl\vert \theta_i\neq 0 \text{ and } \tilde{\theta}_i=0 \text{ or } \tilde{\theta}_i\neq 0 \text{ and } \theta_i=0\}.
$$
In this case, as long as $2s \leq m$, the function is RSS. This implies that the number of data points (samples) should be at least twice the sparsity level. However, in the dense setting (SS), it should equal the number of parameters so that the sufficient condition for strong smoothness holds.
\end{example}
\subsection{Restricted Strong Smoothness for NNs}
If one examines Example \ref{eg:ssandrss} from the perspective of NNs, various interpretations can be drawn. When $n+1 > m$, it is well-known that NNs tend to overfit. Overfitting happens when the model's capacity significantly exceeds the information content of the data, leading the model to fit not only the underlying pattern but also noise and irrelevant details. Ideally, the model should capture only the amount of information that reflects the underlying patterns and regularities. In other words, the objective is to extract the noiseless signal from the dataset using the simplest possible model. This goal is achieved by using a model with reduced complexity, i.e., fewer parameters.
If the overall loss of a less complex model is an RSS function, then finding such a network within a dense structure shown in Fig. (\ref{fig:one-layer}) becomes feasible.
To find such a network, the overall loss function of the network in Fig. \ref{fig:one-layer} must satisfy the conditions in \cite[Corollary 5]{damadi2022gradient}. A key condition in this Corollary is the RSS property. From Example \ref{eg:ssandrss}, it is evident that if $2s \leq m$, the overall loss function qualifies as RSS, meaning $L_{2s}>0$ always exists. Consequently, a network with at most $s$ nonzero parameters can effectively learn from the data.
\subsection{Differentiability is a necessary condition for being RSS}
The overall loss function of a NN is a very complicated scalar-valued function. In order to verify this function is an RSS function, one first needs to make sure that it is a differentiable function. 
The differentiability of the overall loss function, or the overall objective function, depends on the loss function and activation functions used in a NN. The common loss functions like Mean Squared Error, Cross-Entropy, and Binary Cross-Entropy are continuously differentiable over their entire domain. 
Most commonly used activation functions, such as sigmoid and tanh, are differentiable over their entire domain as well. However, some functions, like the ReLU (Rectified Linear Unit) activation function, are not differentiable at every single point; for example, ReLU is not differentiable at 0. Nonetheless, the ReLU function is differentiable almost everywhere, meaning it is not differentiable on a set of measure zero in the domain. Therefore, these points do not affect the overall ability to compute gradients in practice. Consequently, we can assume that the overall loss function of a NN is differentiable.
\subsection{Ensure RSS Property for a NN}\label{sub:learning-rate}

Verifying the RSS property for complex functions is a challenging task. This is because verifying inequality (\ref{eq:rss}) requires considering all sparse vectors with up to $s$ nonzero values in every subspace where the total number of subspaces is ${n \choose s}$. This process must be done for every pair of $\tilde{\bm{\theta}}$ and $\bm{\theta}$ to identify the largest $L_{2s}$ that does not cause a violation for (\ref{eq:rss}). If one ensures that the inequality (\ref{eq:lg}) holds, then the inequality (\ref{eq:rss}) is automatically satisfied. This is because (\ref{eq:lg}) is a sufficient condition for (\ref{eq:rss}). Thus the sufficient condition of equation (\ref{eq:lg}) outlined in Subsection \ref{sub: sufficient} can be a reliable basis for establishing the RSS property.

To ensure that the function maintains the RSS property, we propose the following steps to estimate a conservative upper bound for $L_{2s}$ denoted as $\hat{L}_{2s}$. Essentially, these steps aim to satisfy the inequality (\ref{eq:lg}) by finding a sufficiently large $\hat{L}_{2s}$.

\begin{enumerate}
\item 
Fix a number of monte carlo simulations $n_{\text{monte}}$ and do the following for $n_{\text{monte}}$ times, i.e., for $j \in \{1, \dots, n_{\text{monte}}\}$.
\item 
Randomly initialize a NN with a vector parameter $\bm{\theta}^j \in \mathbb{R}^n$.
\item 
Randomly and without replacement select $s$ nonzero parameters and let the rest be zero.
\item 
Find the smallest nonzero element of the sparse vector parameter in absolute value sense, i.e., 
$$
\delta_s^j = \min_{\theta_i \neq 0, i \in \{1,\dots, n\}} |\theta_i|
$$
\item 
Calculate a perturbed sparse vector as follows:\footnote{If two components are equal in absolute value sense, select the one whose index is smaller. Or, the component can be selected randomly from the equal ones.}
$$
\tilde{\bm{\theta}}^j =H_s(\bm{\theta} + \delta_s^j \mathbf{d}^j)
$$
where $\mathbf{d}^j \in \mathbb{R}^n$ whose components are drawn from a standard normal distribution. 
\item 
Calculate $\hat{L}^j_{2s}$ as follows:
$$
\hat{L}^j_{2s}
=
\frac{\| \nabla f(\tilde{\bm{\theta}}) - \nabla f(\bm{\theta}) \|_2}{\|\tilde{\bm{\theta}}-\bm{\theta}\|_2}.
$$
\item 
Let $\hat{L}_{2s}=\max_{j \in \{1, \dots, n_{\text{monte}}\}} \hat{L}^j_{2s}$ .
\end{enumerate}

\begin{figure}
    \centering
\includegraphics[scale=0.6]{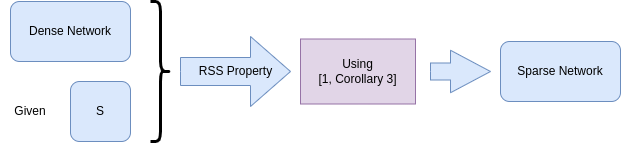}
    \caption{Creating a sparse network with a given $s\leq n-1$.}
    \label{fig:learning-sparse-network}
\end{figure}

Assuming $L_{2s}>0$ of a NN is close to the estimated one ($\hat{L}_{2s}>0$) obtained from the above process, it can be assumed that the underlying neural structure satisfies the RSS property. Thus, the above process, at least with high probability, guarantees satisfaction of the RSS property with $\hat{L}_{2s}>0$. 

By following the steps outlined above, it can be observed that the key assumption (RSS) regarding the overall loss function of a NN (as the objective value of an optimization problem) is satisfied.  Furthermore, the boundedness of every sublevel set over each subspace is always satisfied. This is due to the fact that the loss functions commonly used in NNs, such as Mean Squared Error, Cross-Entropy, and Binary Cross-Entropy, are bounded over their entire domain. 

Building on the aforementioned considerations, the final assumption – that each distinct HT-stable stationary point has distinct objective values – is a more challenging aspect. For the current discussion, we proceed with the assumption that this holds true.

Satisfying all conditions discussed leads to a positive response to Q2. The next aspect to consider is the impact of this estimation on the training process. This consideration is important to Q3, focusing on the practical application of the estimated parameter in training. Details on how this integrates into the overall process will be elaborated in the following subsection.
\subsection{Determining learning rate using estimated $L_{2s}$}
\subsubsection{Non-sparse setting}
The values of $L_{2s}$ and $L$ are crucial in determining the step length (learning rate) for gradient descent-type algorithms, particularly when the objective function is a differentiable and bounded-below function. To elaborate on this fact consider the standard gradient descent scheme outlined in Alg. \ref{alg:GD}. It can be straightforwardly shown that the function value sequence $\big(f(\bm{\theta}^k)\big)$ converges, provided that an arbitrary $0 < \gamma < \frac{2}{L}$ is set, as $k$ goes to infinity.
\subsubsection{Sparse setting}
Fortunately, \cite[Corollary 3]{damadi2022gradient} proves that under the IHT schema outlined in Alg. 1, the function value sequence $\big(f(\bm{\theta}^k)\big)$ converges provided $0 < \gamma \leq \frac{1}{L_{2s}}$.
Hence, by setting the learning rate to be $\gamma =\frac{1}{\hat{L}_{2s}}$, we can set a learning rate for the IHT algorithm which answers Q3. In the Numerical Results section, we will examine Q4 which asks whether the IHT algorithm converges to a sparse local minimizer or not.
The speed or aggressiveness of the IHT algorithm is closely tied to the learning rate, which is inversely proportional to $L_{2s}$, namely the restricted gradient Lipschitz constant of the objective function. Thus, the magnitude of $L_{2s}$ plays a critical role in determining how aggressively the algorithm operates.

\begin{figure}
    \centering
    \includegraphics[width=0.5\linewidth]{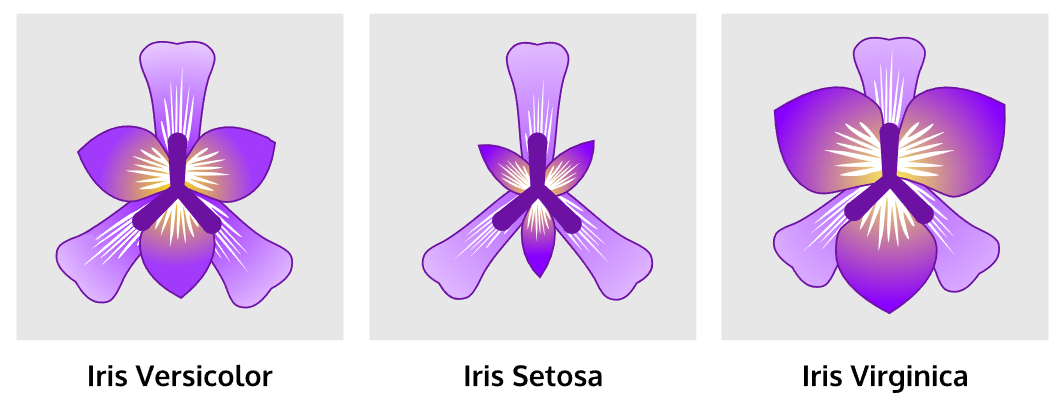}
    \caption[Caption for LOF]{Different variants of iris flowers}
    \label{fig:iris-variants}
\end{figure}

Fig. \ref{fig:learning-sparse-network} provides a schematic representation of the process described, from the starting point of a given dense network and sparsity level 
$s$, through the application of the RSS property, to the resultant sparse network outlined in \cite{damadi2022gradient}.

\begin{table*}[t]
\centering
\caption{Possible number supports for a dense network in Fig. \ref{fig:densevssparse} with 15 parameters}
\label{tab:supports}
\scalebox{1}{
\begin{tabular}{@{}l *{14}{c}@{}}
\hline\toprule
Sparsity levels ($s$) 
 & 1 & 2 & 3 & 4 & 5 & 6 & 7 & 8 & 9 & 10 & 11 & 12 & 13 & 14 
\\\midrule
Possible number supports
 & 15 & 105 & 455 & 1,365 & 3,003 & 5,005 & 6,435 & 6,435 & 5,005 & 3,003 & 1,365 & 455 & 105 & 15
\\
\bottomrule\hline
\end{tabular}
}
\end{table*}

\section{Numerical Results}
After observing the applicability of theoretical results in \cite{damadi2022gradient}, one needs to verify them numerically. This means checking if a smaller part of a NN can learn just as well as the full network. To do so, Pytorch \cite{paszke2019pytorch} is utilized for implementing numerical tests.

\begin{figure}
    \centering
    \includegraphics[scale=0.45]{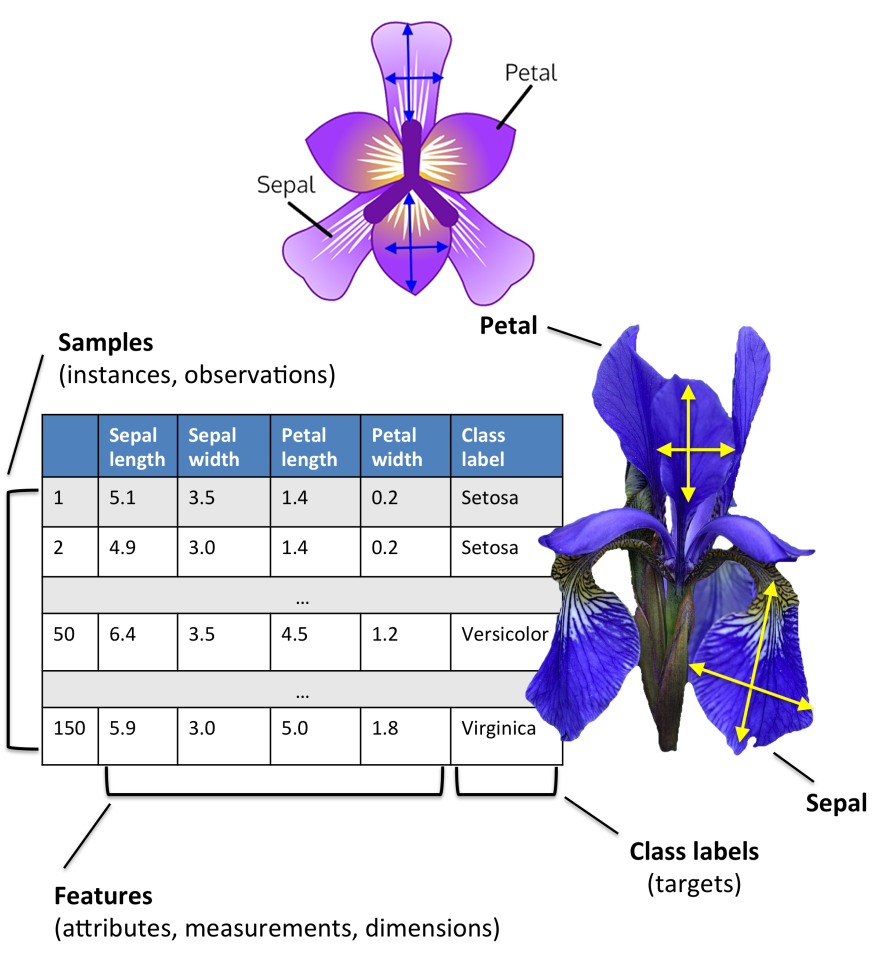}
    \caption{Features representation of the IRIS dataset}
    \label{fig:iris-features}
\end{figure}

To elaborate on the goal, the IRIS dataset is considered  \cite{fisher1936use}.
This dataset has 150 points in $\mathbb{R}^4$.
The dense NN in Fig. \ref{fig:densevssparse} is used as the underlying model for numerical tests. 
This NN processes a four-dimensional vector and classifies it into three different iris flower variants, as shown in Fig. \ref{fig:iris-variants}: Setosa, Versicolour, and Virginica. The various features of a sample are illustrated in Fig. \ref{fig:iris-variants}. Note that all feature values are standardized during the training phase, i.e., each column of Fig. \ref{fig:iris-features} is standardized.

\begin{figure}[ht]
    \centering
    \begin{minipage}{0.5\textwidth}
        \centering
        \includegraphics[scale=0.4]{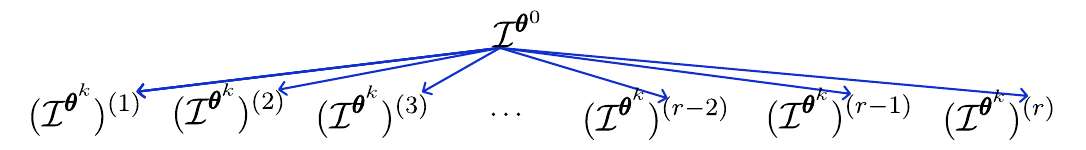}
        \caption{Initialization support vs $k$-th step support}
        \label{fig:initvslast}
    \end{minipage}\hfill
    \begin{minipage}{0.5\textwidth}
        \centering
        \includegraphics[scale=0.45]{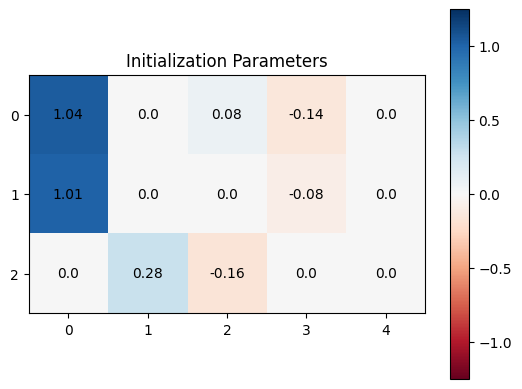}
        \caption{Sparse initialization with random initialization and support seeds of 21 and 84.}
        \label{fig:init-and-supp-seeds}
    \end{minipage}
\end{figure}

Recall the “$\epsilon$-optimality condition” in (\ref{eq:epsopt}). If it holds, there exists a sparse network whose loss is close to that of its dense counterpart.
Suppose after sparsely initializing the NN with a sparsity level of $s=7$, the IHT algorithm converges to a sparse network with only nonzero parameters $w^*_{11}, w^*_{23}, w^*_{31}, w^*_{33}, w^*_{42}, b^*_2,b^*_3$, i.e., the sparse network in Fig. \ref{fig:densevssparse}.
It is not clear which nonzero parameters have been set that resulted in the aforementioned nonzero parameters. In other words, by starting from nonzero parameters $w^0_{12}, w^0_{21}, w^0_{34}, w^0_{33}, w^0_{44}, b^0_1,b^0_3$ as the initialization of our network, there is a chance for ending up with $w^*_{11}, w^*_{23}, w^*_{31}, w^*_{33}, w^*_{42}, b^*_2,b^*_3$ as nonzero parameters. Thus, selecting initial nonzero parameters becomes a hyperparameter. The described situation is shown in Fig. \ref{fig:initvslast}. In this figure, the initialization support $\mathcal{I}^{\bm{\theta}^0}=\text{supp}(\bm{\theta}^0)$ may become any of $r={n \choose s}$ supports at step $k$. The support of a vector is a well-known concept in sparse optimization which is the set of nonzero elements of a vector $\bm{\theta} \in \mathbb{R}^n$. Tab. \ref{tab:supports} shows the number of possible supports for a sparse network. For example, when $s=7$, there are 6,435 possible supports, i.e., $r=6,435$ in Fig. \ref{fig:initvslast}. Hence, in addition to train and test split randomization, and parameter initialization, one has to consider support initialization.
Thus, for each experiment, there are three random seeds: data, initialization, and support seeds.
The data seed is used to split the training and test data randomly by splitting the 150-point IRIS into two random groups of 120 and 30 for training and testing, respectively. In our experiments, this value is referred to as the data seed.
The parameters are initialized randomly by setting the initialization seed, which determines the value of the parameters at initialization. Finally, the support seed determines which parameters are nonzero at initialization.
Fig. \ref{fig:init-and-supp-seeds} shows the values of parameters for a random sparse initialization with initialization and support seeds of 21 and 84.
The trained parameters of this sparse initialization are shown in Fig. \ref{fig:sparse-trained-params}. By comparing the parameters in these two figures, one can clearly observe that the set of non-zero parameters is not the same. This is the power of the IHT algorithm that learns the position of nonzero parameters as well.

\subsection{Validation of “$\epsilon$-optimality condition”}
In order to validate the “$\epsilon$-optimality condition” one needs to train a dense network to obtain the value of $f(\bm{\theta}^*, \mathbf{X}, \mathbf{Y})$ for comparison with $f(\bm{\theta}_s^*, \mathbf{X}, \mathbf{Y})$, where $\bm{\theta}^*$ and $\bm{\theta}_s^*$ are the dense and sparse trained parameters. To do so, 1,000 dense NNs are trained with different random data and initialization seeds. 
For each experiment, randomly splitting the data into 80\%-20\% (120-30 data points) for training and testing, fixes the objective function. This fixed objective function
is run with 10,000 steps and a random $\bm{\theta}^0$ as the initialization that has random nonzero components. The learning rate (step length) is determined using a method outlined in subsection \ref{sub:learning-rate}.

\begin{figure}[ht]
    \centering
    \begin{minipage}{0.5\textwidth}
        \centering
        \includegraphics[scale=0.45]{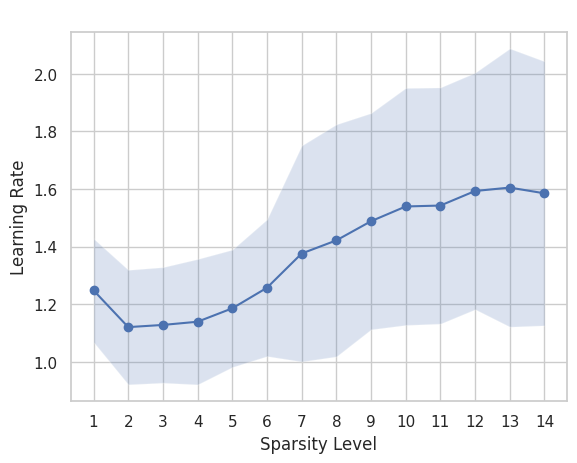}
        \caption{Estimated learning rate for all sparsity levels, i.e., $1 \leq s <n=15$.}
        \label{fig:iris-sparse-learning-rate}
    \end{minipage}\hfill
    \begin{minipage}{0.5\textwidth}
        \centering
        \includegraphics[scale=0.45]{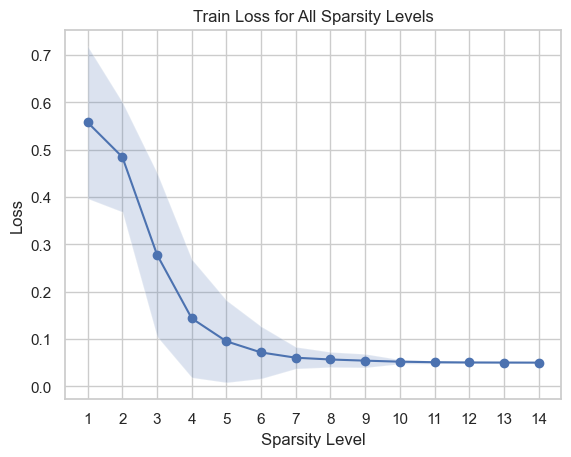}
        \caption{Training loss for all sparsity levels, i.e., $1 \leq s <n=15$.}
        \label{fig:iris-training-loss}
    \end{minipage}
\end{figure}

\begin{figure}[ht]
    \centering
    \begin{minipage}{0.5\textwidth}
        \centering
        \includegraphics[scale=0.45]{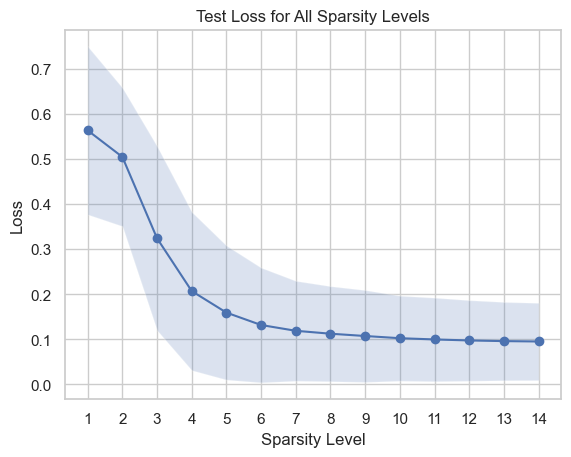}
        \caption{Test loss for all sparsity levels, i.e., $1 \leq s <n=15$.}
        \label{fig:iris-test-loss}
    \end{minipage}\hfill
    \begin{minipage}{0.5\textwidth}
        \centering
        \includegraphics[scale=0.45]{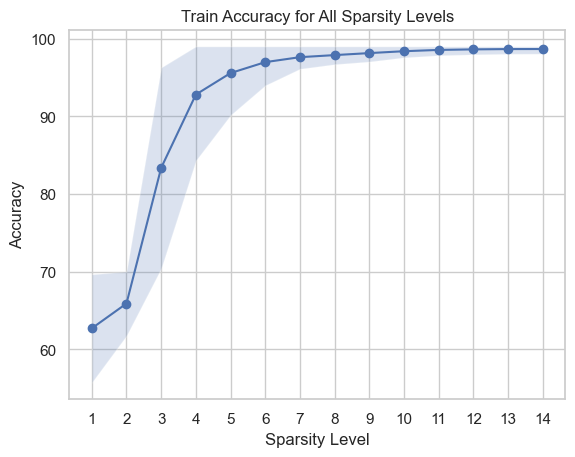}
        \caption{Training accuracy for all sparsity levels, i.e., $1 \leq s <n=15$.}
        \label{fig:iris-training-acc}
    \end{minipage}
\end{figure}

\begin{figure}[ht]
    \centering
    \begin{minipage}{0.5\textwidth}
        \centering
        \includegraphics[scale=0.45]{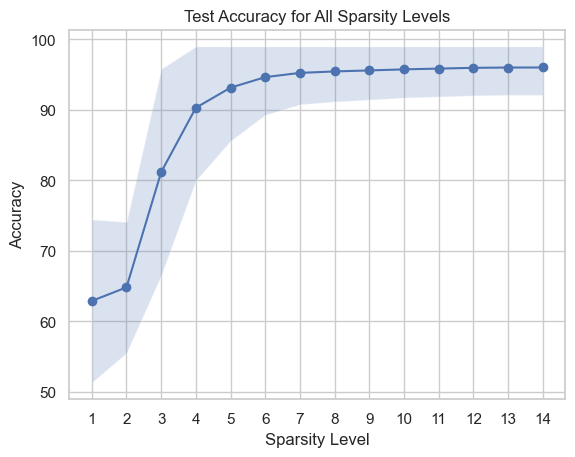}
        \caption{Test accuracy for all sparsity levels, i.e., $1 \leq s <n=15$.}
        \label{fig:iris-test-acc}
    \end{minipage}\hfill
    \begin{minipage}{0.5\textwidth}
        \centering
        \includegraphics[scale=0.45]{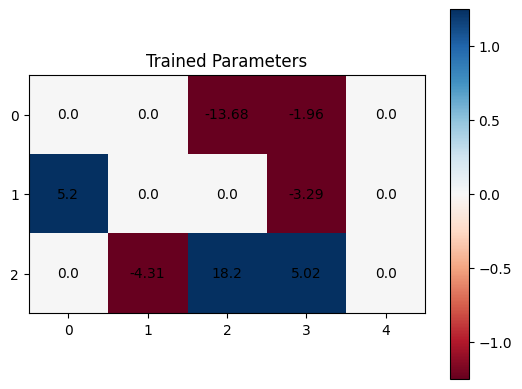}
        \caption{Trained parameters obtained from the IHT with random initialization, data
and support seeds of 21, 42, and 84.}
        \label{fig:sparse-trained-params}
    \end{minipage}
\end{figure}

To find values for $f(\bm{\theta}_s^*, \mathbf{X}, \mathbf{Y})$, one needs to fix the sparsity level. As the dense network has 15 parameters, the sparsity level can range from 1 to 14. Then, 1,000 random experiments are carried out for a fixed sparsity level ranging from $s=1$ to $s=14$. Every experiment is assigned to randomly drawn data, initialization, and support seeds. Once the sparse initialization is determined, one can estimate $\hat{L}_{2s}$ using the process explained in subsection \ref{sub:learning-rate} to set the learning rate. Notice that the learning rate is determined at the beginning of each experiment. Thus the learning rate for each experiment would be different. Fig. (\ref{fig:iris-sparse-learning-rate}) shows how estimated learning rates vary for different $s$ values. As it is illustrated, the learning rates tend to increase as the number of nonzero parameters increases. That means with more parameters, the IHT can go more aggressively.

For $s \geq 10$, the training accuracy in Fig. \ref{fig:iris-training-acc} becomes constant which suggests that there is nothing left that the model can learn.

Fig. \ref{fig:iris-training-loss} shows the training loss variation for different sparsity levels. As this figure shows, the loss value tends to be stabilized around 0.05 because it is the stopping criterion of the IHT algorithm. This numerically verifies that the IHT has converged to a sparse network whose loss is very close to the dense one. Another observation is that having more than 11 nonzero parameters does not make any changes in the training loss values.

As opposed to the training loss, the test loss shown in Fig. \ref{fig:iris-test-loss} has become fixed after a sparsity level of 5. This suggests that the smallest possible sparsity value from the perspective of the test loss is $s=5$. This is an interesting observation because it suggests that only $s=5$ number of parameters required to get a plausible accuracy and increasing $s$ does not change the test loss.

Observing the test accuracy in Fig. \ref{fig:iris-test-acc} reveals that the accuracy becomes more stable after a sparsity level of 10. Particularly noteworthy is the almost constant test accuracy beyond a sparsity level of 5. This implies that as few as 5 nonzero parameters may be sufficient to effectively generalize the IRIS dataset, rather than requiring a larger number of nonzero values.

\begin{figure}
    \centering
    \includegraphics[scale=0.65]{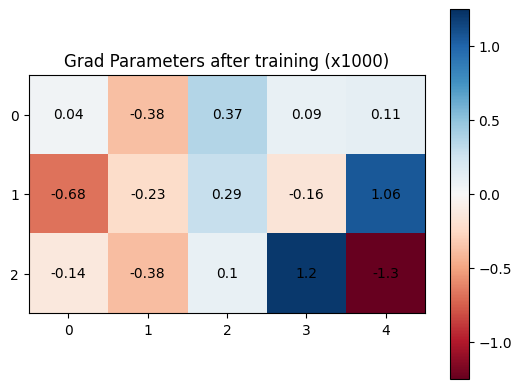}
    \caption{Sparse gradient of trained parameters with random initialization, data and support seeds of 21, 42 and 84.}
    \label{fig:grad-trained-params}
\end{figure}

\subsection{HT-stable points}

Fig. (\ref{fig:sparse-trained-params}) shows the trained parameters of a network with a sparsity level of $s=7$. This network started with a sparse initialization shown in Fig. (\ref{fig:init-and-supp-seeds}). Inspecting the trained model's gradient components in Fig. \ref{fig:grad-trained-params} shows that gradient values are almost zero for all of the parameters. This shows that the IHT algorithm has converged to a HT-stable point because it satisfies the following inequality which is the definition of a HT-stable point in \cite{damadi2022gradient}:

\begin{equation}\label{eq:HTstable}
\min
\Big(
|\theta^*_i|:
i\in \mathcal{S}
\Big)
\ge
\gamma
\max\Big(
|
\nabla_j f(\bm{\theta}^*)
|:
j \notin \mathcal{S}\Big)
\end{equation}
where $\mathcal{S}=\text{supp}(\bm{\theta}^*
)$. This is an affirmative answer to Q4. 
\section{Conclusion}
This paper aims to investigate how recent theoretical progress in sparse optimization can be applied to learning sparse networks. To achieve this, we analyzed the theories in \cite{damadi2022gradient} and identified four main questions: (1) clarifying the basic theoretical assumptions, (2) examining how these assumptions apply to NNs, (3) determining the parameters for these theoretical assumptions, and (4) applying the Iterative Hard Thresholding (IHT) algorithm using these parameters. Our goal is to connect theoretical ideas with their real-world use in learning sparse networks.

\bibliographystyle{IEEEtran}
\bibliography{iclr2023_conference}

\end{document}